%% file: Template.tex
\documentclass{article}
\usepackage{spconf,amsmath,graphicx}
\usepackage{amsmath}    
\usepackage{amssymb}    
\usepackage{amsfonts}   
\usepackage{bm}         

\usepackage[utf8]{inputenc} 
\usepackage{graphicx}     
\usepackage{url}          
\usepackage{booktabs}     
\usepackage{multirow}     
\usepackage{caption}      
\usepackage{subcaption}   
\usepackage{cite}
\usepackage{tikz}
\usepackage{amsmath}
\usepackage{amssymb}
\usetikzlibrary{positioning,shapes,arrows,fit,calc,decorations.markings,decorations.pathreplacing,backgrounds,shadows}

\title{GNN-MoE: Context-Aware Patch Routing using GNNs for Parameter-Efficient Domain Generalization}
\name{Mahmoud Soliman \quad Omar Abdelaziz \quad Ahmed Radwan \quad Anand \quad Mohamed Shehata}
\address{\{mosama97,oabdelaz,ahmedm04,a\}@student.ubc.ca, mohamed.sami.shehata@ubc.ca}

\begin{document}
%
\maketitle
\begin{abstract}
Domain generalization (DG) seeks robust Vision Transformer (ViT) performance on unseen domains. Efficiently adapting pretrained ViTs for DG is challenging; standard fine-tuning is costly and can impair generalization. We propose GNN-MoE, enhancing Parameter-Efficient Fine-Tuning (PEFT) for DG with a Mixture-of-Experts (MoE) framework using efficient Kronecker adapters. Instead of token-based routing, a novel Graph Neural Network (GNN) router (GCN, GAT, SAGE) operates on inter-patch graphs to dynamically assign patches to specialized experts. This context-aware GNN routing leverages inter-patch relationships for better adaptation to domain shifts. GNN-MoE achieves state-of-the-art or competitive DG benchmark performance with high parameter efficiency, highlighting the utility of graph-based contextual routing for robust, lightweight DG.
\end{abstract}

\input{Sections/0_Intro}
\input{Sections/1_RelatedWork}
\input{Sections/2_Methodology}
\input{Sections/3_Exp}
\input{Sections/4_Conclusion}
\bibliographystyle{IEEEtran}
\bibliography{refs}

\end{document}

%% file: Sections/0_Intro.tex
\section{Introduction}
\label{sec:introduction}
Vision Transformers (ViTs) \cite{dosovitskiy2020} excel in computer vision but struggle with domain generalization (DG) \cite{zhou2023}, often overfitting to source domains \cite{Sultana_2022_ACCV} when fully fine-tuned, which is also computationally expensive and prone to catastrophic forgetting. Parameter-Efficient Fine-Tuning (PEFT) methods like Adapters \cite{houlsby2019parameter} and LoRA \cite{Hu2022lora} offer lightweight adaptation by tuning only a small subset of parameters.
Mixture of Experts (MoE) architectures \cite{shazeer2017, fedus2022} extend PEFT by routing inputs to specialized expert sub-networks. However, standard MoE routers typically operate on isolated token features, ignoring inter-patch relationships crucial for robust expert assignment under domain shifts.
To address this, we propose GNN-MoE, a novel framework integrating Graph Neural Networks (GNNs) \cite{kipf2017} for context-aware routing of ViT image patches to highly efficient Kronecker Adapter \cite{he2022parameter} experts. This GNN-based routing on inter-patch graphs enhances adaptation to domain shifts.
Our key contributions are:
\begin{itemize}
\item GNN-MoE: The first framework combining GNN-based routing with parameter-efficient Kronecker Adapters for ViT domain generalization.
\item A graph-based token routing mechanism capturing inter-patch relationships for improved patch-to-expert assignment.
\item State-of-the-art or competitive performance on DG benchmarks with high parameter efficiency.
\end{itemize}

%% file: Sections/1_RelatedWork.tex
\section{Related Work}
\label{sec:related_work}
This section reviews literature relevant to our GNN-MoE framework, covering Domain Generalization, Graph Neural Networks, Parameter-Efficient Fine-Tuning, and Mixture of Experts, highlighting their connection to our approach.
\subsection{Domain Generalization (DG)}
\label{subsec:dg}
Domain Generalization (DG) trains models for robust performance on unseen target domains using data from multiple source domains \cite{zhou2023}, unlike domain adaptation which typically uses target data. The core challenge is learning domain-invariant yet task-relevant representations. DG strategies include domain alignment, meta-learning, learning invariant representations, and data augmentation, underscoring the complexity of out-of-distribution generalization.
\subsection{Graph Neural Networks (GNNs)}
\label{subsec:gnn}
Graph Neural Networks (GNNs) model relationships in graph-structured data via message-passing paradigms. Architectures like Graph Convolutional Networks (GCNs) \cite{kipf2017}, GraphSAGE \cite{hamilton2017graphsage}, and Graph Attention Networks (GATs) \cite{velickovic2018} learn node representations. In vision, GNNs can model contextual relationships between image patches, a principle we leverage for inter-patch routing.
\subsection{Parameter-Efficient Fine-Tuning (PEFT)}
\label{subsec:peft}
Parameter-Efficient Fine-Tuning (PEFT) adapts large pre-trained models like ViTs cost-effectively by freezing most parameters and training only a small subset \cite{houlsby2019parameter}, mitigating catastrophic forgetting and computational demands. PEFT includes adapter modules and Low-Rank Adaptation (LoRA) \cite{Hu2022lora}. Our work uses efficient Kronecker-factorized adapters \cite{he2022parameter}, relevant for DG by enabling domain-specific adaptation while preserving general pre-trained features \cite{dai2022representations}. Kronecker adapters reduce parameter count by decomposing the transformation matrix into compact, structured factors, enabling efficient specialization for each domain expert.
\subsection{Mixture of Experts (MoE)}
\label{subsec:moe}
Mixture of Experts (MoE) architectures enhance model capacity using multiple expert subnetworks and a router for dynamic input token assignment \cite{shazeer2017outrageously}. 
For DG, MoE allows experts (our GNN-MoE uses Kronecker adapters) to specialize in different domain characteristics or domain-invariant features. However, standard MoE routers often ignore inter-patch context, which GNN-MoE addresses.

%% file: Sections/2_Methodology.tex
\section{Methodology}
\label{sec:methodology}

\input{fig/Arch} 
\input{fig/graph_vis} 

The proposed GNN-Routed Mixture-of-Experts Kronecker Adapter (GNN-MoE) module (Figure~\ref{fig:gnn_moe_architecture}) replaces standard linear transformations (e.g., QKV, FFNs) in ViT encoder blocks. Figure~\ref{fig:graph} visualizes a target graph structure.
Input $\bm{X}_{\text{in}} \in \mathbb{R}^{S \times D}$ ($S$ tokens, dimension $D$) is processed via two pathways:

\textbf{Frozen Pathway:} Input $\bm{X}_{\text{in}}$ is processed by the frozen weight matrix $\bm{W}_0 \in \mathbb{R}^{D \times D}$ (snowflake icon, Figure~\ref{fig:gnn_moe_architecture}) yielding $\bm{Y}_{\text{frozen}} = \bm{X}_{\text{in}}\bm{W}_0$. Original bias $\bm{b}_0$, if present, is applied.

\textbf{MoE Adapter Pathway:} Introduces trainable components:
\begin{itemize}
    \item \textbf{GNN Layer (Router):} Processes input $\bm{X}_{\text{in}}$ (or patch subset $\bm{X}_{\text{patches}}$, Sec.~\ref{ssec:gnn_router_orig}) using a trainable GNN on a patch graph to generate routing weights $\bm{G}(\bm{X}_{\text{in}}) \in \mathbb{R}^{S \times N_e}$ for $N_e$ experts.
    \item \textbf{Expert Adapters:} $N_e$ lightweight, trainable Kronecker adapters $\mathcal{E} = \{\text{Adapter}_1, \dots, \text{Adapter}_{N_e}\}$. Each $\text{Adapter}_i$ learns an update matrix $\bm{H}_i \in \mathbb{R}^{D \times D}$ (Sec.~\ref{ssec:kronecker_experts_orig}). Input $\bm{X}_{\text{in}}$ is processed as $\operatorname{Expert}_i(\bm{X}_{\text{in}}) = \bm{X}_{\text{in}}\bm{H}_i$.
    \item \textbf{MoE Combination:} Expert outputs are combined via routing weights $\bm{G}(\bm{X}_{\text{in}})$. For token $s$:
        \begin{equation}
            (\bm{Y}_{\text{MoE}}(\bm{X}_{\text{in}}))_{s,:} = \sum_{i=1}^{N_e} G(\bm{X}_{\text{in}})_{s,i} \cdot (\bm{X}_{\text{in}}\bm{H}_i)_{s,:}.
            \label{eq:moe_combination}
        \end{equation}
\end{itemize}

\textbf{Final Module Output:} Outputs from both pathways are summed with a trainable bias $\bm{b}_{\text{adapter}} \in \mathbb{R}^{D}$:
\begin{equation}
    \bm{X}_{\text{out}} = \bm{X}_{\text{in}}\bm{W}_0 + \sum_{i=1}^{N_e} \bm{G}(\bm{X}_{\text{in}})_{:,i} \odot (\bm{X}_{\text{in}}\bm{H}_i) + \bm{b}_{\text{adapter}}
    \label{eq:module_output_final}
\end{equation}
where $\odot$ is element-wise multiplication with broadcasting of $G(\bm{X}_{\text{in}})_{:,i}$.
The GNN router conditions adapter selection on $\bm{X}_{\text{in}}$'s context. Trainable components (GNN router, Kronecker adapter factors $\bm{H}_i$, $\bm{b}_{\text{adapter}}$) are optimized end-to-end.

\subsection{Kronecker Adapter Experts ($\operatorname{Expert}_i$)}
\label{ssec:kronecker_experts_orig}
Each $\operatorname{Expert}_i$ computes an update matrix $\Delta\bm{W}_i$ via Parameter-efficient Hypercomplex Multiplication (PHM). The effective update matrix $\bm{H}_i \in \mathbb{R}^{D_{\text{in}} \times D_{\text{out}}}$ ($\Delta\bm{W}_i^T = \bm{H}_i$) is:
\begin{equation}
    \bm{H}_i =  \operatorname{Dropout}\left( \sum_{k=1}^{d_{\text{phm}}} \left( (\bm{H}_{\text{shared}})_k \otimes (\bm{H}_{\text{expert\_factors},i})_k \right) \right)
    \label{eq:expert_H_kron_orig}
\end{equation}
where $\otimes$ is Kronecker product. $(\bm{H}_{\text{shared}})_k$ is the $k$-th slice of a shared tensor from PHM rule matrices (e.g., $\bm{P}_L, \bm{P}_R \in \mathbb{R}^{d_{\text{phm}} \times d_{\text{phm}} \times 1}$). $(\bm{H}_{\text{expert\_factors},i})_k$ $\in \mathbb{R}^{(D_{\text{in}}/d_{\text{phm}}) \times (D_{\text{out}}/d_{\text{phm}})}$

is the $k$-th slice from low-rank factors as $\bm{L}_{i} \bm{R}_{i}$ ($d_{\text{phm}}$ is PHM dimension, $r_i$ is expert rank). Summation is over PHM slices $k$. Expert output:
\begin{equation}
    \operatorname{Expert}_i(\bm{X}) = \bm{X} \bm{H}_i.
    \label{eq:expert_output_final_corrected_orig}
\end{equation}

\subsection{GNN-Based Router}
\label{ssec:gnn_router_orig}
A GNN-based router determines weights $\bm{G}(\bm{X})$ for MoE, operating on a graph from patch tokens $\bm{X}_{\operatorname{patches}} \in \mathbb{R}^{N_p \times D_{\text{in}}}$ (excluding class token $\bm{z}_{\operatorname{cls}}$).

\textbf{Graph Construction Strategies:} Defines connectivity $\operatorname{EdgeIndex} \in \mathbb{Z}^{2 \times N_{\text{edges}}}$.
\textbf{\textit{Spatial Adjacency:}} Connects immediate spatial neighbors (e.g., 8-connectivity) for local context.
\textbf{\textit{Radius:}} Connects patches $v_j, v_k$ if Euclidean distance $\|\operatorname{coord}(v_j) - \operatorname{coord}(v_k)\|_2 \le r_{\text{th}}$ (Eq.~\ref{eq:radius_graph_orig}), found most effective.
\begin{equation}
    (v_j, v_k) \in \mathcal{E} \iff \|\operatorname{coord}(v_j) - \operatorname{coord}(v_k)\|_2 \le r_{\text{th}}.
    \label{eq:radius_graph_orig}
\end{equation}
\textbf{\textit{Fully Connected:}} Connects all patch nodes for global context (higher cost). Self-loops are added for all types. Graph structure influences available context.

\textbf{GNN Architecture:} We use GNNs for contextualized representations. Primarily Graph Convolutional Network (GCN)~\cite{kipf2017}. Given initial patch features $\bm{h}_v^{(0)} = (\bm{X}_{\operatorname{patches}})_v$, GCN layer $l$ updates $\bm{h}_v^{(l)}$:
\begin{equation}
    \bm{h}_v^{(l)} = \sigma\left(\sum_{u \in \mathcal{N}(v) \cup \{v\}} \frac{1}{\sqrt{\deg(v)\deg(u)}} \bm{W}^{(l)} \bm{h}_u^{(l-1)}\right). \label{eq:gcn_update_revised}
\end{equation}
$\mathcal{N}(v)$ are neighbors of $v$, $\deg(v)$ is degree, $\bm{W}^{(l)}$ is learnable weight, $\sigma$ is activation. A stack of $L_{\operatorname{GNN}}$ GCN layers computes final embeddings $\bm{H}_{\operatorname{patches}}^{\operatorname{context}}$.
Ablations (Sec.~\ref{sec:exp}) explored GraphSAGE~\cite{hamilton2017graphsage} and GAT~\cite{velickovic2018}. Layer Normalization is applied between GNN layers if $L_{\operatorname{GNN}} > 1$.

\textbf{Routing Weights Generation:} GNN output $\bm{H}_{\operatorname{patches}}^{\operatorname{context}}$ generates routing weights. An $\operatorname{MLP}_{\operatorname{router}}$ projects embeddings to $\operatorname{Scores}_{\operatorname{patches}} \in \mathbb{R}^{N_p \times N_e}$:
\begin{equation}
    \operatorname{Scores}_{\operatorname{patches}} = \operatorname{MLP}_{\operatorname{router}}(\bm{H}_{\operatorname{patches}}^{\operatorname{context}}).
    \label{eq:router_scores_orig}
\end{equation}
Scores are converted to probabilities $\operatorname{W}_{\operatorname{patches}}$ via softmax, with optional noise $\bm{\epsilon}_{\text{noise}} \sim \mathcal{N}(0, (\text{gate\_noise}/N_e)^2 \bm{I})$ and temperature $\tau$:
\begin{equation}
    \operatorname{W}_{\operatorname{patches}} = \operatorname{Softmax}\left((\operatorname{Scores}_{\operatorname{patches}} + \bm{\epsilon}_{\operatorname{noise}}) / \tau\right).
    \label{eq:router_weights_patch_orig}
\end{equation}
Class token weights $\operatorname{W}_{\operatorname{cls}} \in \mathbb{R}^{1 \times N_e}$ (e.g., by averaging $\operatorname{W}_{\operatorname{patches}}$):
\begin{equation}
    (\operatorname{W}_{\operatorname{cls}})_i = \frac{1}{N_p} \sum_{j=1}^{N_p} (\operatorname{W}_{\operatorname{patches}})_{j,i}.
    \label{eq:router_weights_cls_orig}
\end{equation}
Routing matrix $\bm{G}(\bm{X}) = \operatorname{CONCAT}(\operatorname{W}_{\operatorname{cls}}, \operatorname{W}_{\operatorname{patches}})$.

\subsection{Training Objective}
The composite loss (Eq.~\ref{eq:total_loss_final_orig}):
\begin{equation}
    \mathcal{L}_{\operatorname{total}} = \mathcal{L}_{\operatorname{task}} + \lambda_{\operatorname{aux}} \cdot \mathcal{L}_{\operatorname{aux}}.
    \label{eq:total_loss_final_orig}
\end{equation}
$\mathcal{L}_{\operatorname{task}}$ is classification cross-entropy; $\mathcal{L}_{\operatorname{aux}}$ is load balancing loss~\cite{shazeer2017outrageously}. If $P_{s,i} = G(\bm{X})_{s,i}$ and $N_T = B \cdot S$ (batch size $B$):
\begin{equation}
    \mathcal{L}_{\operatorname{aux}} = \frac{N_e}{N_T^2} \sum_{i=1}^{N_e} \left( \sum_{s=1}^{N_T} P_{s,i} \right)^2.
    \label{eq:aux_loss_moe_orig}
\end{equation}
$\lambda_{\operatorname{aux}}$ balances terms.

%% file: fig/Arch.tex
\begin{figure}[t]
\centering
\resizebox{\columnwidth}{!}{
\begin{tikzpicture}[
    mainbox/.style={draw, thick, minimum width=4.5cm, minimum height=1.8cm, align=center, rounded corners=3pt, drop shadow},
    expertbox/.style={draw, thick, minimum width=3.2cm, minimum height=1.6cm, fill=violet!15, align=center, rounded corners=3pt, drop shadow},
    frozenbox/.style={draw, thick, minimum width=3.8cm, minimum height=1.6cm, fill=blue!10, align=center, rounded corners=3pt, drop shadow},
    routerbox/.style={draw, thick, minimum width=5cm, minimum height=3.5cm, fill=orange!15, align=center, rounded corners=3pt, drop shadow},
    arrow/.style={->, >=stealth, thick, color=black!80},
    frozenarrow/.style={->, >=stealth, thick, color=blue!70, line width=1.5pt},
    expertarrow/.style={->, >=stealth, thick, color=violet!70, line width=1.5pt},
    routerarrow/.style={->, >=stealth, thick, color=orange!70, line width=1.5pt},
    matrixlabel/.style={font=\small, midway, fill=white, inner sep=2pt, rounded corners=1pt},
    dimensionlabel/.style={font=\footnotesize, color=black!70},
    ]
    
    \node[mainbox, fill=green!10] (input) at (0,-6) {
        \textbf{Input Features}\\
        $\mathbf{X}_{\text{in}} \in \mathbb{R}^{B \times N \times D}$\\
        {\footnotesize (Batch, Nodes, Features)}
    };
    
    \node[frozenbox] (backbone) at (-5,-2) {
        \textbf{Frozen Backbone}\\
        $\mathbf{W}_0 \in \mathbb{R}^{D \times D}$\\
        {\footnotesize Pretrained weights}
    };
    
    \node[routerbox] (router) at (0,-2) {
        \textbf{GNN Router} $\mathcal{R}$\\[0.3em]
        \begin{tikzpicture}[scale=0.8, transform shape]
            \foreach \i in {1,2,3,4,5} {
                \node[circle, draw, fill=orange!30, minimum size=0.6cm] (n\i) at ({72*(\i-1)}:1.2) {};
            }
            \foreach \i in {1,2,3,4,5} {
                \foreach \j in {1,2,3,4,5} {
                    \ifnum\i<\j
                        \draw[->, >=stealth, color=orange!70] (n\i) -- (n\j);
                    \fi
                }
            }
        \end{tikzpicture}\\[0.2em]
        {\footnotesize Outputs: $\mathbf{G}(\mathbf{X}) \in \mathbb{R}^{B \times N \times N_e}$}
    };
    
    \node[expertbox] (expert1) at (7.5,1) {
        \textbf{Expert}$_1$\\
        $\mathbf{A}_1 \in \mathbb{R}^{D \times D}$
    };
    \node[expertbox] (expert2) at (7.5,-1) {
        \textbf{Expert}$_2$\\
        $\mathbf{A}_2 \in \mathbb{R}^{D \times D}$
    };
    \node[expertbox] (expert3) at (7.5,-3) {
        \textbf{Expert}$_3$\\
        $\mathbf{A}_3 \in \mathbb{R}^{D \times D}$
    };
    \node[font=\Large, color=violet!70] (vdots) at (7.5,-3) {$\vdots$};
    \node[expertbox] (expertN) at (7.5,-5) {
        \textbf{Expert}$_{N_e}$\\
        $\mathbf{A}_{N_e} \in \mathbb{R}^{D \times D}$
    };
    
    \node[draw, dashed, rounded corners=5pt, fit=(expert1) (expert2) (expert3) (vdots) (expertN), 
          inner sep=12pt, color=violet!70, line width=1.5pt] (experts_container) {};
    \node[above=5pt of experts_container, font=\small\bfseries, color=violet!70] {
        Domain-Specific Expert Set $\mathcal{E}$
    };
    
    \node[mainbox, fill=red!10] (aggregation) at (0,2) {
        \textbf{Weighted Aggregation}\\
        $\bigoplus$ (Element-wise Sum)\\
        {\footnotesize Routing-weighted combination}
    };
    
    \node[mainbox, fill=green!20] (output) at (0,5) {
        \textbf{Output Features}\\
        $\mathbf{X}_{\text{out}} \in \mathbb{R}^{B \times N \times D}$\\
        {\footnotesize Domain-adapted features}
    };
    
    \draw[frozenarrow] (input.west) -- ++(-2.73,0) -- (backbone.south)
        node[matrixlabel, pos=0.3] {$\mathbf{X}_{\text{in}}$};
    
    \draw[routerarrow] (input) -- (router)
        node[matrixlabel, right, xshift=0.5cm] {$\mathbf{X}_{\text{in}}$};
    
    
    \draw[routerarrow] (router.east) -- ++(0.5,0) |- (experts_container.west)
        node[matrixlabel, pos=0.6] {$\mathbf{G}(\mathbf{X})$};
    
    \draw[frozenarrow] (backbone.north) -- ++(0,0) |- (aggregation.west)
        node[matrixlabel, pos=0.7, above] {$\mathbf{X}_{\text{in}} \mathbf{W}_0$};
    
    \draw[expertarrow] (expert1.west) -- ++(-1.4,0) |- (aggregation.east)
        node[matrixlabel, pos=0.7, above] {$\sum_i \mathbf{G}_i \odot (\mathbf{X} \mathbf{A}_i)$};
    \draw[expertarrow] (expert2.west) -- ++(-1.2,0) |- (aggregation.east);
    \draw[expertarrow] (expert3.west) -- ++(-1,0) |- (aggregation.east);
    \draw[expertarrow] (expertN.west) -- ++(-0.8,0) |- (aggregation.east);
    
    \draw[arrow] (aggregation) -- (output)
        node[matrixlabel, left] {$\mathbf{X}_{\text{out}}$};
    
    
    \node[above=2pt of router, font=\small, color=orange!70] {{\includegraphics[width=14.25pt,height=17.18pt]{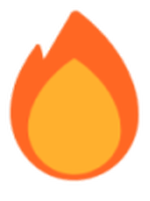}} \textsc{Trainable}};
    \node[right=-20pt of expert1, font=\small, color=violet!70] {{\includegraphics[width=14.25pt,height=17.18pt]{fig/fire.png}}};
    \node[right=-20pt of expert2, font=\small, color=violet!70] {{\includegraphics[width=14.25pt,height=17.18pt]{fig/fire.png}}};
    \node[right=-20pt of expert3, font=\small, color=violet!70] {{\includegraphics[width=14.25pt,height=17.18pt]{fig/fire.png}}};
    \node[right=-20pt of expertN, font=\small, color=violet!70] {{\includegraphics[width=14.25pt,height=17.18pt]{fig/fire.png}}};
    
    
    
\end{tikzpicture}}
\caption{GNN-Routed Mixture-of-Experts Architecture for Domain Generalization. The architecture combines a frozen pretrained backbone with trainable GNN-based routing and domain-specific expert adapters. The GNN router analyzes input structure to generate routing weights, enabling adaptive combination of domain experts for robust cross-domain performance.}
\label{fig:gnn_moe_architecture}
\end{figure}

%% file: fig/graph_vis.tex
\begin{figure}
    \centering
    \resizebox{1\columnwidth}{!}{
        \includegraphics[width=0.5\linewidth]{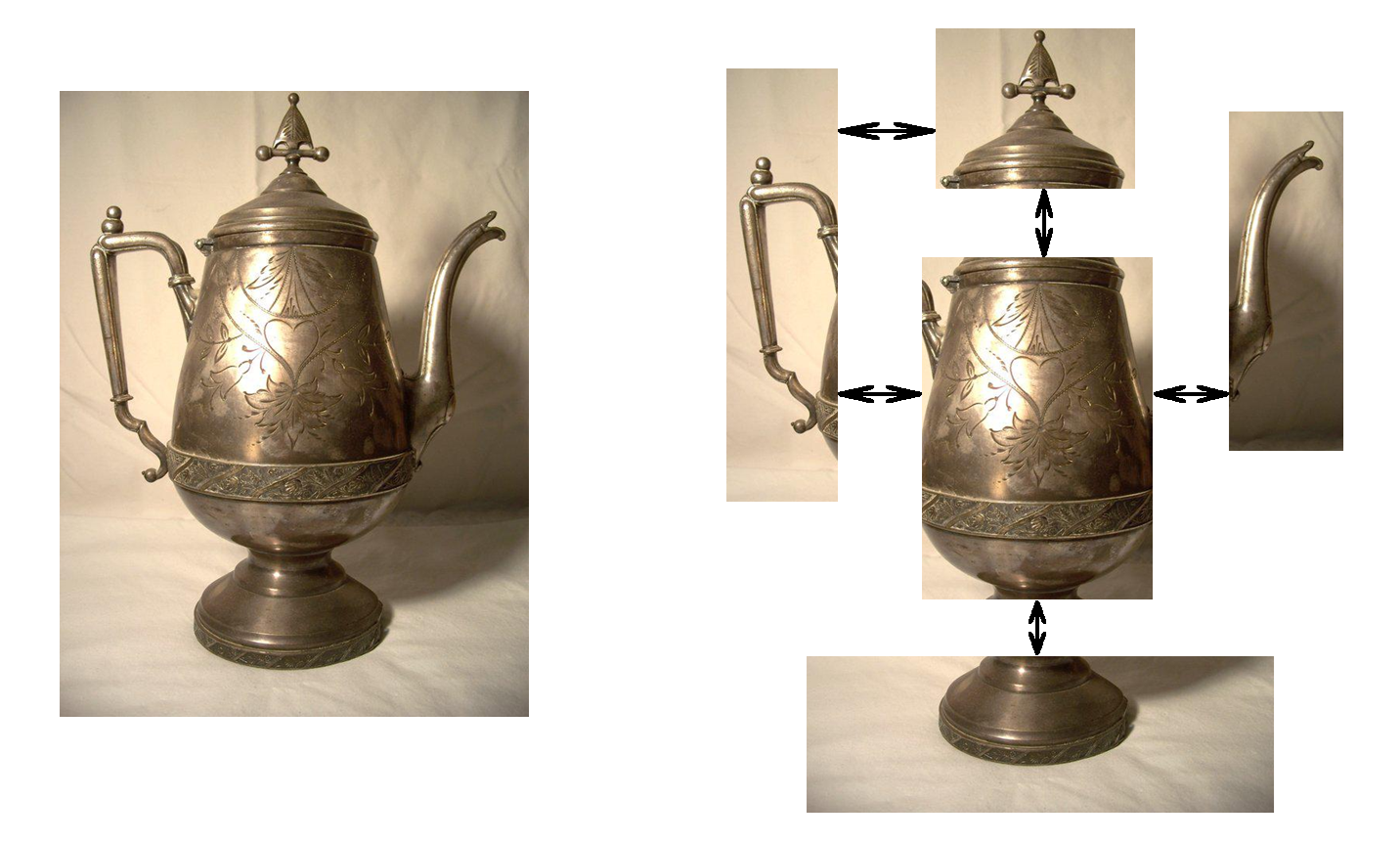}
    }
    \caption{A representation of an old kettle of OfficeHome dataset broken down into graph nodes. An optimal GNN router should be capable of understanding this relational graph in order to route patches to the corresponding experts.}
    \label{fig:graph}
\end{figure}

%% file: Sections/3_Exp.tex
\section{Experiments}
\label{sec:exp}

\begin{table*}[htbp]
\centering
\caption{Comparison of different domain generalization methods.}
\label{tab:domain_generalization_main}
\resizebox{\textwidth}{!}{
\begin{tabular}{@{}lllrrrrrrrr@{}}
\toprule
Algorithm & Architecture & Pretraining & PACS & VLCS & OfficeHome & TerraIn. & DomainNet & Avg. & \#Param. & Trainable \#Param. \\
\midrule

MIRO & ViT-B/16 & CLIP & 96.7 $\pm$ 0.7 & 82.4 $\pm$ 0.3 & 87.3 $\pm$ 0.5 & 52.3 $\pm$ 0.5 & 50.6 $\pm$ 0.6 & 73.9 & 172M & 85.8M \\

ZS-CLIP & ViT-B/16 & CLIP & 96.1 $\pm$ 0.0 & 82.3 $\pm$ 0.0 & 81.8 $\pm$ 0.0 & 33.8 $\pm$ 0.0 & 56.6 $\pm$ 0.0 & 70.2 & 85.8M & 85.8M \\

Lin. Prob. & ViT-B/16 & CLIP & 94.9 $\pm$ 1.4 & 77.5 $\pm$ 0.7 & 79.3 $\pm$ 0.2 & 44.6 $\pm$ 2.1 & 48.2 $\pm$ 0.2 & 68.9 & 85.8M & 85.8M \\
CoOp & ViT-B/16 & CLIP & 96.4 $\pm$ 0.3 & 80.8 $\pm$ 0.3 & 83.0 $\pm$ 0.1 & 46.8 $\pm$ 0.7 & 59.5 $\pm$ 0.2 & 73.6 & 85.8M & 85.8M \\
CoCoOp & ViT-B/16 & CLIP & 96.7 $\pm$ 0.2 & 80.3 $\pm$ 0.3 & 83.4 $\pm$ 0.2 & 45.3 $\pm$ 2.4 & 59.4 $\pm$ 0.2 & 73.2 & 85.8M & 85.8M \\
DPL & ViT-B/16 & CLIP & 96.4 $\pm$ 0.3 & 80.9 $\pm$ 0.5 & 83.0 $\pm$ 0.3 & 46.6 $\pm$ 0.8 & 59.5 $\pm$ 0.3 & 73.6 & 85.8M & 85.8M \\
VP & ViT-B/16 & CLIP & 95.8 $\pm$ 0.1 & 82.2 $\pm$ 0.0 & 81.2 $\pm$ 0.2 & 34.9 $\pm$ 0.2 & 56.5 $\pm$ 0.0 & 70.1 & 85.8M & 85.8M \\
VPT & ViT-B/16 & CLIP & 96.9 $\pm$ 0.2 & 82.0 $\pm$ 0.2 & 83.2 $\pm$ 0.1 & 46.7 $\pm$ 0.6 & 58.5 $\pm$ 0.2 & 73.6 & 85.8M & 85.8M \\
MaPLe & ViT-B/16 & CLIP & 96.5 $\pm$ 0.2 & 82.2 $\pm$ 0.2 & 83.4 $\pm$ 0.0 & 50.2 $\pm$ 0.9 & 59.5 $\pm$ 0.3 & 74.4 & 89.3M & 89.3M  \\
SPG & ViT-B/16 & CLIP & 97.0 $\pm$ 0.5 & 82.4 $\pm$ 0.4 & 83.6 $\pm$ 0.4 & 50.2 $\pm$ 1.2 & 60.1 $\pm$ 0.5 & 74.7 & 85.8M & 85.8M \\

PromptStyler & ViT-B/16 & CLIP & 97.2 $\pm$ 0.1 & 82.9 $\pm$ 0.0 & 83.6 $\pm$ 0.0 & - & 59.4 $\pm$ 0.0 & - & 85.8M & 85.8M \\
PromptStyler & ViT-L/14 & CLIP & 98.6 $\pm$ 0.0 & 82.4 $\pm$ 0.2 & 89.1 $\pm$ 0.0 & - & 65.5 $\pm$ 0.0 & - & 307M & 307M \\
ERM$^{\dagger}$ & RegNetY-16GF & SWAG$_{\text{IG3B}}$ & 89.6 $\pm$ 0.4 & 78.6 $\pm$ 0.3 & 71.9 $\pm$ 0.6 & 51.4 $\pm$ 1.8 & 48.5 $\pm$ 0.6 & 68.0 & 83.6M & 83.6M \\
MIRO$^{\dagger}$ & RegNetY-16GF & SWAG$_{\text{IG3B}}$ & 97.4 $\pm$ 0.2 & 79.9 $\pm$ 0.6 & 80.4 $\pm$ 0.2 & 58.9 $\pm$ 1.3 & 53.8 $\pm$ 0.1 & 74.1 & 167.2M & 83.6M \\
GMDG & RegNetY-16GF & SWAG$_{\text{IG3B}}$ & 97.3 $\pm$ 0.1 & 82.4 $\pm$ 0.6 & 80.8 $\pm$ 0.6 & \textbf{60.7 $\pm$ 1.8} & 54.6 $\pm$ 0.1 & 75.1 & 83.6M & 83.6M \\

\midrule
\multicolumn{11}{@{}l@{}}{\textit{Methods using additional supervision}} \\
{VL2V-ADiP} & {ViT-B/16} & {CLIP} & {94.9} & {81.9} & {85.7} & {55.4} & {59.4} & {75.5} & {235.8M} & {83.6M} \\
{VL2V-SD} & {ViT-B/16} & {CLIP} & {96.7} & {83.3} & {87.4} & {58.5} & {62.8} & {77.7} & {235.8M} & {83.6M} \\
\midrule
\multicolumn{11}{@{}l@{}}{\textit{Methods with Parameter-Efficient Fine-Tuning (PEFT)}} \\
 ERM (Baseline) & ViT-B/16 & CLIP & 85.8 $\pm$ 2.1 & 78.5 $\pm$ 0.9 & 78.1 $\pm$ 0.8 & 41.0 $\pm$ 1.6 & 52.2 $\pm$ 0.1 & 67.1 & 85.8M & 85.8M \\
 ERM$_{\text{Compacter}}$ & ViT-B/16 & CLIP & 94.1 $\pm$ 0.4 & 81.0 $\pm$ 0.5 & 83.0 $\pm$ 0.1 & 35.9 $\pm$ 0.7 & 56.2 $\pm$ 1.2 & 70.0 & 85.9M & \textbf{0.10M} \\
 ERM$_{\text{Attention}}$ & ViT-B/16 & CLIP & 93.8 $\pm$ 0.6 & 82.0 $\pm$ 0.3 & 85.9 $\pm$ 0.4 & 51.4 $\pm$ 0.8 & 57.2 $\pm$ 0.1 & 74.1 & 85.9M & 28.4M \\
 ERM$_{\text{LoRA, r=2}}$ & ViT-B/16 & CLIP & 96.4 $\pm$ 0.6 & 82.6 $\pm$ 0.6 & 86.7 $\pm$ 0.3 & 46.1 $\pm$ 1.7 & 61.5 $\pm$ 0.1 & 74.7 & 85.9M & \textbf{0.11M} \\
 ERM$_{\text{KAdaptation}}$ & ViT-B/16 & CLIP & 97.5 $\pm$ 0.1 & 83.0 $\pm$ 0.1 & 90.3 $\pm$ 0.1 & 51.9 $\pm$ 0.5 & \textbf{62.7 $\pm$ 0.0} & 77.1 & 85.9M & \textbf{0.14M} \\
\midrule
\multicolumn{11}{@{}l@{}}{\textit{Methods with Mixture-of-Adapter (MoA)}} \\
 ERM$_{\text{LoRA-MoA}}$ & ViT-B/16 & CLIP & 96.9 $\pm$ 0.3 & 82.8 $\pm$ 0.7 & 89.5 $\pm$ 0.2 & 49.2 $\pm$ 2.4 & 62.2 $\pm$ 0.0 & 75.9 & 87.2M & 1.5M \\ 
 ERM$_{\text{KAdaptation-MoA}}$ & ViT-B/16 & CLIP & 97.4 $\pm$ 0.2 & 83.1 $\pm$ 0.3 & 90.6 $\pm$ 0.0 & 52.8 $\pm$ 1.4 & \textbf{62.7 $\pm$ 0.1} & 77.3 & 87.3M & 1.5M \\ 
 ERM$_{\text{GNN-MoE}}$ & ViT-B/16 & CLIP & \textbf{97.85 $\pm$ 0.1} & \textbf{83.6 $\pm$ 0.1} & \textbf{90.7 $\pm$ 0.1} & 53.8 $\pm$ 0.1 & 62.3 $\pm$ 0.1 & \textbf{77.7} & 87.6M & 1.8M \\ 
\bottomrule
\end{tabular}%
} 
\end{table*}

We empirically evaluate our GNN-MoE framework for Domain Generalization (DG), comparing its accuracy and parameter efficiency against state-of-the-art (SOTA) methods on standard DG benchmarks and validating designs via ablation studies.

\subsection{Experimental Setup}

\subsubsection{Datasets}
\label{sssec:datasets}
GNN-MoE is evaluated on five DG benchmarks: \textbf{PACS}~\cite{Li2017deeper} (7 categories), \textbf{VLCS}~\cite{Torralba2011unbiased} (5 categories), \textbf{OfficeHome}~\cite{Venkateswara2017deep} (65 categories), \textbf{TerraIncognita}~\cite{Beery2018recognition} (10 categories), and \textbf{DomainNet}~\cite{Peng2019moment} (345 categories), all exhibiting diverse distribution shifts. We use the standard leave-one-domain-out protocol (3 random seeds, mean $\pm$ std. dev.).

\subsubsection{Implementation Details}
\label{sssec:implementation_details}
Experiments use ViT-B/16 (CLIP LAION-2B pretrained~\cite{radford2021learning}). GNN-MoE modules replace QKV projections in alternating frozen encoder blocks. Only GNN-MoE modules and classification head are trained. Main results: $N_e=4$ Kronecker adapter experts (ranks $\bm{r}=[1,2,4,8]$, $d_{\text{phm}}=128$). AdamW (LR $10^{-4}$, WD $10^{-5}$), $\lambda_{\text{aux}}=0.01$, 8 epochs, batch 32. Implemented in PyTorch/PyTorch Geometric~\cite{fey2019fastgraph}.

\subsection{Comparison with Domain Generalization Methods}
\label{ssec:comparison_sota_main_revised}

GNN-MoE (GCN Architecture) is compared against existing DG methods (Table~\ref{tab:domain_generalization_main}).

\subsubsection{Baselines (Full Fine-Tuning \& Standard PEFT)}
GNN-MoE (77.7\% avg. accuracy) substantially outperforms full ViT-B/16 fine-tuning ERM (67.1\%) and standard PEFT (ERM$_{\text{KAdaptation}}$, 77.1\%). This highlights the benefit of GNN-MoE's structure and context-aware routing for parameter-efficient DG.

\subsubsection{Advanced Domain Generalization Methods \& MoA}
GNN-MoE (77.7\% avg. accuracy) also surpasses advanced DG methods like MIRO~\cite{Wortsman2022robust} (73.9\%) and MoA approaches like ERM$_{\text{KAdaptation-MoA}}$~\cite{lee2024mixtureofadapters} (77.3\%), achieving SOTA or competitive results on PACS (97.85\%), VLCS (83.6\%), OfficeHome (90.7\%), TerraIncognita (53.8\%), and DomainNet (62.3\%).

\subsection{Ablation Studies}
\label{sec:ablation_studies}
To understand the impact of different architectural choices and hyperparameters, we conducted a series of ablation studies.
Unless otherwise specified, experiments are conducted on the OfficeHome dataset. We select strong performing configurations as baselines and vary one component at a time where possible.

\subsubsection{Impact of GNN Architecture}
\label{sssec:ablation_gnn_type}
We investigate how different GNN backbones affect performance. We use a configuration with 128 hidden channels, 1 layer, 0.1 dropout, and a Radius graph (mean aggregation, full on OfficeHome as the baseline (Table~\ref{tab:ablation_gnn_type_officehome}).

\begin{table}[htbp]
\centering
\caption{Ablation on GNN Type (OfficeHome) full graph.}
\label{tab:ablation_gnn_type_officehome}
\resizebox{0.4\columnwidth}{!}{
\begin{tabular}{lllcr}
\toprule
GNN Type & Result (\%) \\
\midrule
GCN & 90.70 \\
SAGE  & 90.50 \\
GAT  & 90.20 \\
GATV2 & 89.93 \\
\bottomrule
\end{tabular}}
\end{table}
On OfficeHome with these parameters, GCN performs slightly better than SAGE and GAT, with GATV2 showing a minor decrease. The differences are relatively small, suggesting robustness across these GNN types for this particular setup.



\subsubsection{Impact of Graph Construction Method}
\label{sssec:ablation_graph_type}

We compare Radius, full, and spatial graphs for SAGE and GCN on OfficeHome, keeping other parameters (128 hidden, 1 layer, 0.1 dropout) constant (Table~\ref{tab:ablation_graph_type_dynamic_static}).
\begin{table}[htbp]
\centering
\caption{Ablation on Graph Types (OfficeHome).}
\label{tab:ablation_graph_type_dynamic_static}
\resizebox{0.8\columnwidth}{!}{
\begin{tabular}{lllcr}
\toprule
GNN Type & Graph Type & Radius & Result (\%) \\
\midrule
SAGE & Radius & 2.9 & 90.70 \\
SAGE & Full   & N/A & 90.50 \\
SAGE & Spatial   & N/A & 90.40 \\
\midrule
GCN & Radius  & 2.9 & 90.23 \\
GCN & Full    & N/A & 90.70 \\
GCN & Spatial    & N/A & 90.46 \\
\bottomrule
\end{tabular}
}
\end{table}
For SAGE, the Radius graph (mean aggregation) slightly outperforms the full graph. Conversely, for GCN, the full graph yields a better result than the specific Radius graph configuration tested. This suggests the optimal graph construction method can be GNN-dependent.


\subsubsection{Impact of Radius Value}
\label{sssec:ablation_radius_value}
We assess the sensitivity to the `Radius` parameter for SAGE with a Radius graph (mean aggregation) on OfficeHome, using 128 hidden channels, 1 layer, and 0.1 dropout (Table~\ref{tab:ablation_radius_value}).
\begin{table}[htbp]
\centering
\caption{Ablation on Radius Value (OfficeHome, SAGE).}
\label{tab:ablation_radius_value}
\begin{tabular}{cr}
\toprule
Radius & Result (\%) \\
\midrule
1.5    & 90.50 \\
2.8    & 90.70 \\
2.9    & 90.70 \\
3.2    & 90.60 \\
4.5    & 90.45 \\
\bottomrule
\end{tabular}
\end{table}
The model performance is relatively stable for radius values between 2.8 and 3.5 for this configuration, with a slight peak around 2.8-2.9.

\subsubsection{Impact of Model Capacity and Regularization}
\label{sssec:ablation_capacity_regularization}

\begin{table}[htbp]
    \centering
    \caption{SAGE Hyperparameters (OfficeHome, Radius Graph, $R_val$ as per baseline).}
    \label{tab:ablation_sage_hyper}
    \begin{tabular}{@{}llr@{}}
    \toprule
    Parameter        & Value & Result (\%) \\
    \midrule
    Hidden Channels  & 64    & 90.54 \\
                     & 128   & 90.70 \\
                     & 256   & 90.26 \\
    \midrule
    Num. GNN Layers  & 1     & 90.61 \\ 
                     & 2     & 90.25 \\
    \midrule
    Dropout Rate     & 0.1   & 90.61 \\ 
                     & 0.2   & 90.61 \\
                     & 0.3   & 90.60 \\
                     & 0.5   & 90.37 \\
                     
    \bottomrule
    \end{tabular}%
\end{table}

In Table~\ref{tab:ablation_sage_hyper} we examine the effect of varying hidden channels for SAGE with a Radius graph (mean aggregation, R=2.9) on OfficeHome (1 layer, 0.1 dropout). Increasing hidden channels from 128 to 256 leads to a decrease in performance. Data for 64 channels with strictly comparable parameters was not available.
Next, we compare 1 vs. 2 layers for SAGE with a Radius graph (mean aggregation, R=3.5) on OfficeHome (128 hidden, 0.1 dropout). Increasing the number of layers from 1 to 2 results in a performance drop for this specific configuration.
Finally, we assess the effect of dropout for SAGE with a Radius graph (mean aggregation, R=3.5) on OfficeHome (128 hidden, 1 layer). For this SAGE configuration, changing dropout from 0.1 to 0.3 has a negligible impact on performance.


\subsection{Visual Results}
\input{fig/gnn_tsne}
We show that the learned feature embeddings by our model align in the t-SNE space such that each class embedding is more separable than a baseline CLIP \cite{radford2021learning} model in Figure~\ref{fig:gnn_tsne}. 

%% file: fig/gnn_tsne.tex
\begin{figure}
    \centering
    \resizebox{1\columnwidth}{!}{
        \includegraphics[width=0.5\linewidth]{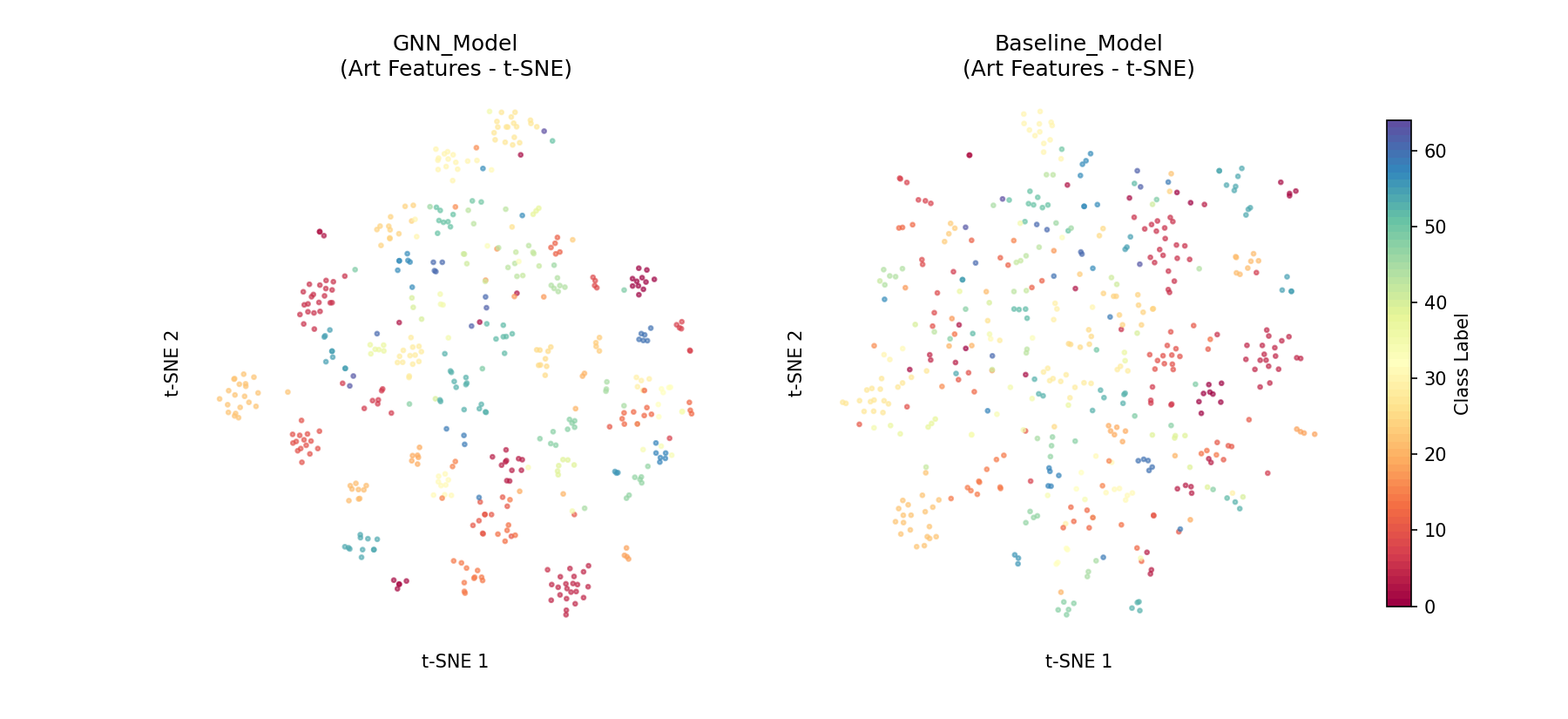}
    }
    \caption{A t-SNE projection of the learned features of our model versus a baseline model on the ART domain of OfficeHome. Best viewed when zoomed in.}
    \label{fig:gnn_tsne}
\end{figure}

%% file: Sections/4_Conclusion.tex
\section{Conclusion}
\label{sec:Conclusion}
This work introduced GNN-MoE, a parameter-efficient framework for Vision Transformer domain generalization. By using a GNN-based router to contextually assign image patches to specialized Kronecker adapter experts, GNN-MoE achieves state-of-the-art or competitive performance on DG benchmarks. This demonstrates the value of graph-based relational reasoning for efficient and robust adaptation in domain-shifted scenarios. It can likely be integrated into other Vision Transformer variants, such as ViT-Large or Swin Transformers, to enhance their adaptability and generalization capabilities.